\documentclass{article}
\usepackage[final]{neurips_2019}
\usepackage[utf8]{inputenc}
\usepackage[T1]{fontenc}    
\usepackage{hyperref}    
\usepackage{url}         
\usepackage{booktabs}       
\usepackage{amsfonts}      
\usepackage{nicefrac}     
\usepackage{microtype}      
\usepackage{svg}
\usepackage{amsmath}
\usepackage{arydshln} 
\usepackage{tabularx}
\usepackage[export]{adjustbox}  
\usepackage{longtable}
\newcolumntype{K}{>{\raggedright\arraybackslash}m}

\title{Towards better healthcare:\\ What could and should be automated?}  

\author{%
  Wolfgang Frühwirt \\
  Machine Learning Research Group \\
  University of Oxford, UK  \\
  \And
  Paul Duckworth\thanks{corresponding: paul.duckworth@eng.ox.ac.uk} \\
  Machine Learning Research Group \\
  University of Oxford, UK \\
  }

\begin{document}

\maketitle

\begin{abstract}

While artificial intelligence (AI) and other automation technologies might lead to enormous progress in healthcare, they may also have undesired consequences for people working in the field. In this interdisciplinary study, we capture empirical evidence of
not only what healthcare work \textit{could} be automated, but also what \textit{should} be automated. We quantitatively investigate these research questions by utilizing probabilistic machine learning models trained on thousands of ratings, provided by both healthcare practitioners and automation experts. Based on our findings, we present an analytical tool (Automatability-Desirability Matrix) to support policymakers and organizational leaders in developing practical strategies on how to harness the positive power of automation technologies, while accompanying change and empowering stakeholders in a participatory fashion. 
\end{abstract}

\section{Introduction}
From improving diagnostic quality~\citep{Gulshan2019} to enhancing accuracy of treatment outcome predictions~\citep{Koutsouleris2016}, artificial intelligence (AI) is already having life-changing impact in healthcare. 
As AI technologies become more ubiquitous, their effects are likely to amplify and have a large economic impact. According to recent estimates, healthcare is one of the sectors with the largest economic potential of applied AI ~\citep{Rao2017}.

At the same time, there is strong public concern that AI and other technologies enabling automation will lead to undesirable substitution of human labour. In other words, while automation technologies might lead to enormous progress in the quality and efficiency of healthcare products and services, they may have undesired consequences for people 
in the field~\citep{frey2019technology}. Therefore, in the present study, we not only address the question \textit{``What could be automated?''}, but also \textit{``What should be automated?''}. 

Previous studies have investigated the automatability of entire occupations~\citep{Frey-Osb2017}, and their constituent work activities~\citep{Duckworth2019,Manyika2017future}
from the viewpoint of technical experts from science and industry. To the best of our knowledge, our study is the first to make automatability predictions based on empirical data from domain experts, i.e., practitioners actually working in the field.  Spending a large portion of their lifetime performing the examined activities and being members of their specialized scientific and professional communities, healthcare practitioners have domain-specific insights into the automatability of their work that technical experts don't. Conversely, the latter have technical insight in the general automatability of work that domain experts don't. Therefore, we utilize predictive machine learning models for both groups and compare perspectives. 

As a secondary contribution, and an equally important research question, we introduce an empirical investigation of the \textit{desirability} of automation of activities in the domain of research.
We directly survey healthcare professionals, i.e. the people affected, on how desirable the full automation of individual work activities would be,
and similarly build a predictive model over all activities performed by healthcare professionals.

Based on our findings, we present an analytical tool, Automatability-Desirability Matrix (AD Matrix), to support policymakers and organizational leaders in  developing practical strategies on how to harness the positive power of AI and other automation technologies, while accompanying change and empowering stakeholders in a participatory fashion.

\section{Data and Methods}
\paragraph{Survey Data}
We recruited 150 practitioners from a variety of healthcare professions. The ethics committee of the University of Oxford approved the study and all participants gave informed consent. An online questionnaire presented healthcare experts with activities routinely performed by somebody of their profession and asked their opinion on how \textit{automatable} these activities are today (using currently available technology), and how \textit{desirable} the complete automation of these activities would be.

The technical experts analyzed were participants of the study by~\citet{Duckworth2019}, where 156 academic and industry experts in machine learning, robotics, intelligent systems, and operations research were surveyed. The online questionnaire used for surveying healthcare experts differed insofar as we directly asked for work activity ratings whereas \citet{Duckworth2019} asked for task ratings and indirectly inferred activity scores. This novelty potentially improves the reliability of our data and therefore the quality of the resulting prediction models. 

Both groups of experts answered the following question: \textit{``Do you believe that technology exists today that could automate these tasks?''} by labeling each activity (healthcare experts) or task (technical experts) as either: \textit{Not automatable today} (score of 1.0), \textit{Mostly not automatable today (human does most of it)} (2.0), \textit{Could be mostly automated today (human still needed)} (3.0), \textit{Completely automatable today} (4.0), or \textit{Unsure}. 
Domain experts were asked to label all work activities of their particular occupation, whereas technical experts were presented with five occupations and their five “most important” constituent tasks.

Healthcare experts were additionally asked how desirable the complete automation of their work activities would be: \textit{``For you, how desirable would it be, if the following activities could be completely automated?''}: \textit{Very undesirable} (1.0), \textit{Undesirable} (2.0), \textit{Desirable} (3.0), \textit{Very desirable} (4.0), \textit{Unsure}. 

Classification of occupations, work activities, and tasks were all derived from the O*NET database ~\citep{NationalCenterforO*NETDevelopment}, as provided by the US Department of Labor.

\paragraph{Machine Learning Model with Uncertainty}
We seek a flexible function estimation capable of modeling complex, non-linear relationships between features of work, and the automatability and desirability of that work. Given that we are modeling subjective human preferences we also desire a measure of model uncertainty. The key idea, as described in~\citet{Duckworth2019}, is to transform the occupational-level characteristics $\boldsymbol{x}_{o}$ relating to 35 skills, 33 knowledge areas and 52 abilities required to perform an occupation, into representative feature vectors for each work activity, $\boldsymbol{x}_{w} = \big[ \boldsymbol{x}^{s}_{o}, \boldsymbol{x}^{k}_{o}, \boldsymbol{x}^{a}_{o} \big] \in \mathbb{R}^{+120}$, in the O*NET database. 

The surveyed healthcare professionals provided 2\,608 ratings of how \textit{automatable} their work activities are, and 2\,278 ratings of how desirable full automation would be. We combined each activity's multiple domain expert labels using Independent Bayesian Classifier Combination (IBCC), a principled Bayesian approach to combine multiple classifications~\citep{kim2012bayesian,simpson2013dynamic}. 
IBCC creates a posterior prediction over class labels that reflects the individual labelers' tendencies to agree with others over ultimately chosen label values. 

In line with~\citet{Duckworth2019} we trained a Gaussian Process (GP) with an ordinal likelihood function~\citep{chu2005gaussian} using GPFlow~\citep{GPflow2017} to reflect the nature of having discrete labels with an ordinal interpretation on uncertain data. 
A kernel density estimate of the expert ordinal data for both automatability (blue) and desirability of automation (orange) is presented in Figure~\ref{fig:results} (left). 

\section{Results \& Discussion}

We used the ordinal GP models to infer the \textit{automatability} and \textit{desirability} of automation for all healthcare work activities (N $=191$)~\footnote{We restricted our test set to those work activities performed by healthcare occupations (see Appendix B).}. We allowed the GP posterior mean to correct/update our experts' opinions, inspired by similar work at the occupational level by~\citet{Frey-Osb2017}. Table~\ref{table-results} reports the three work activities with the highest and lowest automatability (top) and desirability of automation scores (bottom) inferred by our probabilistic model. For an extended table see Appendix~A. 
\begin{center}
\begin{table}[h!]
\small
\caption{Three largest/smallest inferred work activity automatability (top) and desirability of automation (bottom) scores, sorted by the output of the domain expert model (uncertainties are shown in parenthesis).
}\label{table-results}
\vskip 0.05in
\begin{tabular}{@{}p{8.5cm}cc@{}}
\hspace{3cm}Work Activity                                             & Healthcare Experts & Technical Experts \\ \midrule
Automatability                                            &                    &                   \\ \midrule
Enter patient or treatment data into computers.           & 2.57 (0.46)        & 3.21 (0.60)       \\
Process medical billing information.                      & 2.57 (0.49)        & 3.31 (0.61)       \\
Supervise medical support personnel.                      & 2.56 (0.47)        & 2.72 (0.61)       \\ \cdashline{1-3}
Advise others on healthcare matters.                      & 1.84 (0.41)        & 1.51 (0.68)       \\
Counsel clients on mental health or personal achievement. & 1.75 (0.40)        & 1.38 (0.70)       \\
Diagnose neural or psychological disorders.               & 1.75 (0.41)        & 2.06 (0.77)       \\ \midrule

\vspace{0.5em}
Desirability of Automation                                &                    &                   \\ \midrule
Maintain medical equipment or instruments.                & 3.74 (0.42)        &                  \\
Write reports or evaluations.                             & 3.71 (0.42)        &                  \\
Instruct patients in the use of assistive equipment.      & 3.64 (0.59)        &                  \\ \cdashline{1-3}
Counsel clients on mental health or personal achievement. & 1.65 (0.36)        &                  \\
Treat dental problems or diseases.                        & 1.61 (0.36)        &                 \\
Examine mouth, teeth, gums, or related facial structures. & 1.56 (0.37)        &                 \\ \midrule
\end{tabular}
\end{table}
\end{center}

\vspace{-2em}
\paragraph{Automatability of Work Activities}
Overall, we found that technical experts are more optimistic than healthcare domain experts regarding the current potential for automatability of work
($\mu, \sigma = (2.54, 0.47)$ vs $(2.26, 0.18)$: $p <$ 0.0001). The predictions of the domain experts' model had a significantly lower level of predictive uncertainty than that of technical experts 
($\mu, \sigma = (0.45, 0.25)$ vs $(0.65, 0.005)$: $p <$ 0.0001). 
Perhaps this reflects the higher inter-rater reliability of healthcare professionals, or it might be contributed to the fact that we had better label coverage around the inferred scores of the smaller test set.

\paragraph{Desirability of Automation}
With a mean prediction of 2.83 ($\sigma = 0.44$), our machine learning model indicates that healthcare practitioners are open to the full automation of many of their occupational activities (see Table~\ref{table-results} (bottom) or Appendix A for an extended table). However, as we present in the next section, in many instances, there is a difference between what healthcare professionals perceive could and should be automated. 

Desirability of automating work activities is found to be significantly correlated with automatability itself ($r$: 0.29, $p$: $<$ 0.0001); both inferred from domain experts' models. 
This small-sized effect might indicate a bias regarding personal preferences in automatability ratings or a general tendency that non-automatable work is more desirable. 

\paragraph{Automatability-Desirability Matrix}
\begin{figure}[h!]
  \centering
  \includegraphics[width=0.48\linewidth, valign=t]{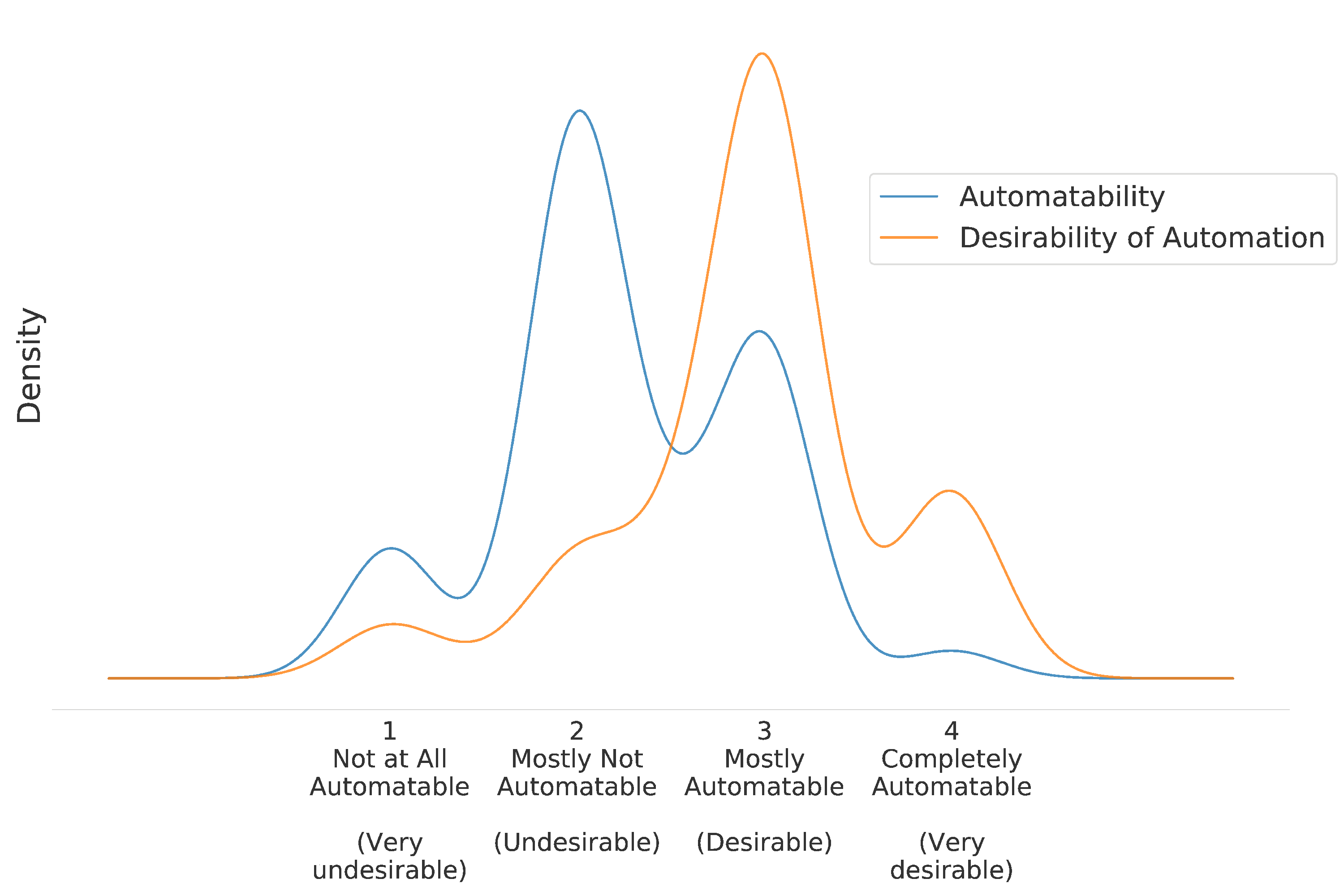} \hspace{1em}
  \includegraphics[width=0.48\linewidth, valign=t]{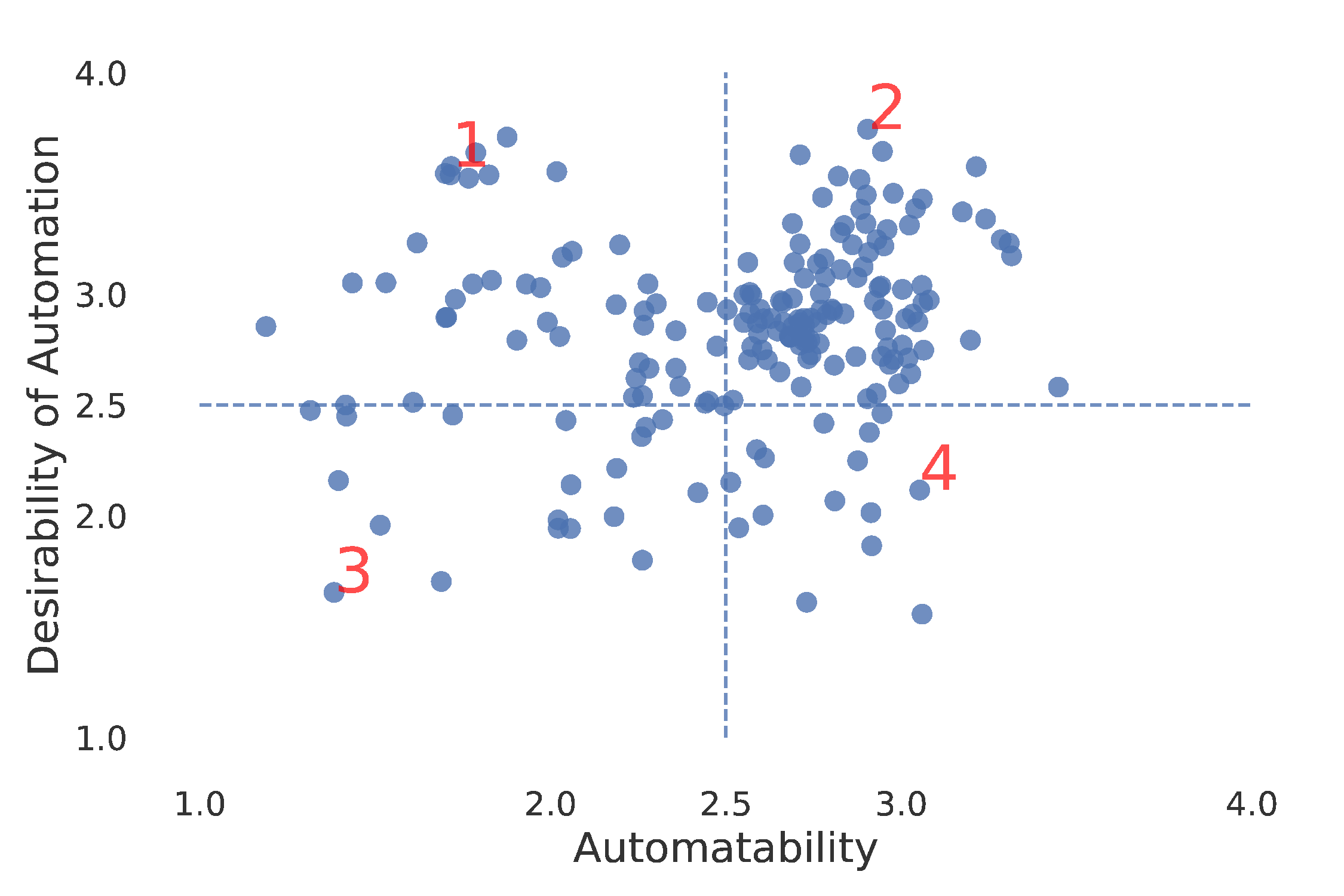}
  \caption{
  Left: Kernel Density Estimate of healthcare expert ordinal data for automatability (blue) and desirabilty of automation (orange). %
  Right: Automatability-Desirability Matrix, including an exemplar in each quadrant (see text for details).}
  \label{fig:results}
\end{figure}

To support governments and non-governmental organizations in assessing current conditions and developing future strategies, we present the AD Matrix. We combine the automatability of work activities, inferred from the technical experts’ predictive model, with the desirability of automation of those activities from the domain practitioners’ model. Figure~\ref{fig:results} (right) presents the AD Matrix. 

To demonstrate the actionable insights that can be gained, we highlight particularly interesting example work activities in each of the four quadrants. 
1. ``Develop treatment plans for patients or clients'' is an activity considered highly desirable to automate by practitioners, however current technology is still lacking.
2. ``Maintain medical equipment or instruments'' is considered highly desirable to automate and also technically possible to automate. 
3. ``Counsel clients on mental health or personal achievement'' has a low desirability to automate and also remains technically very challenging to do so. 
4. ``Operate on patients to treat conditions'' is considered an activity that can be technically automated, however there is little desire from healthcare practitioners to. 

\paragraph{Limitations and Social Impact}
One limitation of our work is that we only considered the opinions of healthcare professionals regarding the desirability of automating work activities. In doing so, we make healthcare practitioners our primary stakeholders. Whereas other groups, e.g. patients, might have different preferences. 
We do so, as we believe practitioners spend a large part of their lifetime performing the work activities represented by the O*NET database and are central to the healthcare system. 
Furthermore, recent research suggests increased automation in primary healthcare is not expected to result in job losses, but instead necessary to tackle an ever increasing quantity of work~\citep{willis2019future}. 
Given the potential to alleviate healthcare workers from rote activities, we aim to capture their preferences to achieve better job satisfaction, which is so crucial for healthcare quality. 

Secondly, we intend the AD Matrix to be used as a tool to support strategic decision making. However, as interdisciplinary researchers ourselves, we would like to stress that the findings of our work should be included in a wide range of considerations when making decisions that impact the lives of various stakeholders. Crucially, the outcomes presented in this study should not be used to drive automation in a way that is not beneficial for society in general. 

\section{Conclusion}

Our study concludes that technical experts are more optimistic than healthcare domain experts regarding the current potential for automatability of healthcare work. 
We also observe that healthcare professionals desire a surprisingly high proportion of work activities to be fully automated. 

Furthermore, we provide a tool considering human workforce preferences, the AD Matrix, for identifying work activity automatability and desirability of automation in a succinct four quadrant model that can be used for strategic guidance. We hope to have contributed to a more accurate understanding of the challenging field of healthcare automation,
considering its stakeholders, with the ultimate goal of better healthcare outcomes.

\medskip
\small
\newpage
\bibliography{bibfile}
\bibliographystyle{nips}

\newpage

\section*{Appendix A: Predicted Work Activity Scores}


\begin{table}[h!]
\centering
\small
\caption{All inferred automatability scores and corresponding desirability of automation scores, sorted by automatability from the healthcare model. Compared to technical experts from~\citet{Duckworth2019}. }\label{tab:apps:auto}
\begin{tabular}{@{}p{8.5cm}ccc@{}}
\toprule
 \multicolumn{3}{r}{\textbf{Healthcare Experts}} \hspace{3em}  & \textbf{Tech} \\ 
\textbf{Work Activity} & \textbf{Automatability} &  \textbf{Desirability} & \textbf{Experts} \\
\midrule



                                         Diagnose neural or psychological disorders. &    1.75  (0.4) &  2.14  (0.5) &  2.06  (0.8) \\
                                          Counsel clients on mental health or personal achievement. &    1.75  (0.4) &  1.65  (0.4) &  1.38  (0.7) \\
                                                                Advise others on healthcare matters. &    1.84  (0.4) &  1.96  (0.4) &  1.51  (0.7) \\
                                                              Advise others on educational matters. &    1.85  (0.4) &  2.48  (0.5) &  1.32  (0.7) \\
                              Design psychological or educational treatment procedures or programs. &    1.85  (0.4) &  2.16  (0.4) &   1.4  (0.7) \\
                                                      Direct medical science or healthcare programs. &    1.89  (0.4) &  2.79  (0.5) &   1.9  (0.7) \\
                                            Administer standardized physical or psychological tests. &     1.9  (0.4) &  2.51  (0.4) &  1.61  (0.7) \\
                                 Attend conferences or workshops to maintain professional knowledge. &    1.93  (0.4) &  2.95  (0.4) &  2.19  (0.7) \\
                                Conduct scientific research of organizational behavior or processes. &    1.93  (0.4) &  2.85  (0.5) &  1.19  (0.7) \\
                                                                                 Supervise trainees. &    1.94  (0.4) &   1.7  (0.4) &  1.69  (0.7) \\
                                                                      Schedule medical facility use. &    1.96  (0.4) &  2.55  (0.5) &  2.93  (0.7) \\
                                                               Establish standards for medical care. &    1.96  (0.4) &  2.86  (0.6) &  2.27  (0.7) \\
                                              Counsel clients or patients regarding personal issues. &    1.97  (0.4) &  2.45  (0.4) &  1.42  (0.6) \\
                                                     Counsel clients regarding interpersonal issues. &    1.97  (0.4) &  2.45  (0.4) &  1.72  (0.6) \\
                                                                         Diagnose dental conditions. &    1.99  (0.4) &  2.01  (0.5) &  2.91  (0.9) \\
                                                         Order medical diagnostic or clinical tests. &     2.0  (0.4) &  2.83  (0.4) &  2.65  (0.6) \\
                                              Select medical equipment for addressing patient needs. &     2.0  (0.4) &   2.1  (0.6) &  2.42  (0.7) \\
                                                                   Lead classes or community events. &    2.01  (0.5) &   2.5  (0.5) &  1.42  (0.7) \\
            Advise athletes, coaches, or trainers on exercise regimens, nutrition, or equipment use. &    2.01  (0.5) &  1.95  (0.7) &  2.54  (0.7) \\
                                                                  Collect information about clients. &    2.01  (0.4) &  3.23  (0.4) &  1.62  (0.6) \\
                                                                              Prescribe medications. &    2.03  (0.4) &  2.73  (0.3) &  2.74  (0.6) \\
                                                            Establish nursing policies or standards. &    2.05  (0.4) &  2.51  (0.6) &  2.44  (0.7) \\
                                                                Direct healthcare delivery programs. &    2.05  (0.4) &  2.77  (0.4) &  2.47  (0.6) \\
                                                     Care for women during pregnancy and childbirth. &    2.06  (0.4) &  2.85  (0.4) &  2.69  (0.7) \\
                                                     Advise patients on healthcare system processes. &    2.06  (0.4) &   2.7  (0.5) &  2.62  (0.7) \\
                                              Advise medical personnel regarding healthcare issues. &    2.07  (0.4) &  2.89  (0.4) &  2.72  (0.6) \\

                    Analyze quantitative data to determine effectiveness of treatments or therapies. &    2.08  (0.4) &  2.67  (0.4) &  2.28  (0.6) \\
                                                     Develop methods of social or economic research. &    2.08  (0.4) &  3.05  (0.4) &  1.43  (0.7) \\
                                            Counsel clients or patients with substance abuse issues. &    2.08  (0.4) &  3.54  (0.4) &   1.7  (0.6) \\
                                                                                  Immunize patients. &     2.1  (0.4) &  2.81  (0.4) &  2.68  (0.6) \\
                                                            Operate on patients to treat conditions. &     2.1  (0.4) &  2.12  (0.4) &  3.05  (0.7) \\
                                              Evaluate treatment options to guide medical decisions. &     2.1  (0.4) &   2.7  (0.5) &  2.57  (0.6) \\
                                         Confer with clients to discuss treatment plans or progress. &     2.1  (0.4) &  3.03  (0.4) &  1.97  (0.6) \\
                                                                  Prescribe treatments or therapies. &     2.1  (0.4) &  2.79  (0.4) &  2.72  (0.6) \\
                                                Assess patient work, living, or social environments. &    2.11  (0.5) &  2.52  (0.7) &  2.52  (0.7) \\
                                                                    Develop medical treatment plans. &    2.11  (0.4) &  2.77  (0.4) &  2.71  (0.6) \\
                Interview clients to gather information about their backgrounds, needs, or progress. &    2.11  (0.4) &  3.52  (0.4) &  1.77  (0.6) \\
                                                      Treat patients using psychological therapies. &    2.11  (0.4) &  1.94  (0.4) &  2.02  (0.7) \\
                                  Evaluate the effectiveness of counseling or educational programs. &    2.12  (0.4) &  3.06  (0.4) &  1.83  (0.6) \\
                                                                        Diagnose medical conditions. &    2.12  (0.4) &  2.87  (0.3) &  2.66  (0.6) \\
                                                                          Establish treatment goals. &    2.12  (0.5) &   1.8  (0.5) &  2.26  (0.8) \\
                     Collaborate with other professionals to assess client needs or plan treatments. &    2.12  (0.4) &  3.54  (0.4) &  1.71  (0.6) \\
                                                    Develop treatment plans for patients or clients. &    2.12  (0.4) &  3.58  (0.4) &  1.72  (0.6) \\
                                      Advise patients on effects of health conditions or treatments. &    2.13  (0.4) &  2.89  (0.4) &  2.63  (0.6) \\
                                      Analyze test data or images to inform diagnosis or treatment. &    2.13  (0.4) &  2.86  (0.3) &  2.72  (0.6) \\
                                      Teach life skills or strategies to clients or their families. &    2.13  (0.4) &  3.05  (0.4) &  1.93  (0.6) \\
                                             Supervise workers providing client or patient services. &    2.13  (0.4) &  3.05  (0.4) &  1.78  (0.6) \\

\bottomrule
\end{tabular}
\end{table}                                                       
                                                                
\newpage   
Appendix A Table (continued) 
\begin{table}[h!]
\centering
\small
\begin{tabular}{@{}p{8.5cm}ccc@{}}
\toprule
 \multicolumn{3}{r}{\textbf{Healthcare Experts}} \hspace{3em}  & \textbf{Tech} \\ 
\textbf{Work Activity} & \textbf{Automatability} &  \textbf{Desirability} & \textbf{Experts} \\
\midrule

                                                       Advise patients on preventive care techniques. &    2.13  (0.4) &  2.68  (0.4) &  2.97  (0.7) \\
                                                      Help clients get needed services or resources. &    2.13  (0.4) &  3.19  (0.4) &  2.06  (0.6) \\
                                                                      Plan social sciences research. &    2.14  (0.5) &  2.89  (0.5) &   1.7  (0.7) \\
                                                     Administer blood or other fluids intravenously. &    2.14  (0.4) &  2.26  (0.5) &  2.61  (0.7) \\
                                                                Treat chronic diseases or disorders. &    2.14  (0.4) &  2.78  (0.3) &  2.73  (0.6) \\                                                       
                             Instruct college students in social sciences or humanities disciplines. &    2.14  (0.5) &  3.05  (0.8) &  1.53  (0.8) \\
                              Refer patients to other healthcare practitioners or health resources. &    2.14  (0.4) &  2.87  (0.3) &  2.76  (0.6) \\
                                                      Advise others on social or educational issues. &    2.14  (0.4) &  2.87  (0.4) &  1.99  (0.6) \\
                                                  Adjust dental devices or appliances to ensure fit. &    2.15  (0.5) &   2.0  (0.4) &  2.61  (0.8) \\
                                                                            Maintain client records. &    2.15  (0.4) &  3.54  (0.4) &  1.82  (0.6) \\
                                          Prescribe assistive medical devices or related treatments. &    2.15  (0.4) &  2.59  (0.5) &  2.99  (0.7) \\
                                                Administer anesthetics or sedatives to control pain. &    2.15  (0.4) &  1.86  (0.4) &  2.92  (0.6) \\
                                                     Treat acute illnesses, infections, or injuries. &    2.16  (0.4) &  2.25  (0.4) &  2.88  (0.6) \\
                                                  Analyze medical data to determine cause of death. &    2.16  (0.5) &   2.5  (0.8) &  2.49  (0.8) \\
                                                     Monitor clients to evaluate treatment progress. &    2.16  (0.4) &  3.64  (0.4) &  1.79  (0.6) \\
                                          Monitor patients following surgeries or other treatments. &    2.16  (0.4) &  2.64  (0.5) &  3.03  (0.7) \\
                             Communicate detailed medical information to patients or family members. &    2.16  (0.4) &  2.93  (0.4) &  2.77  (0.6) \\
                                              Examine patients to assess general physical condition. &    2.16  (0.4) &  2.71  (0.3) &  2.73  (0.6) \\
                                                          Design public or employee health programs. &    2.16  (0.4) &  2.96  (0.4) &   2.3  (0.6) \\
                                                          Develop exercise or conditioning programs. &    2.16  (0.5) &  2.15  (0.8) &  2.51  (0.7) \\
                                                                      Develop educational programs. &    2.17  (0.4) &   2.9  (0.4) &   1.7  (0.6) \\
                                                                  Supervise patient care personnel. &    2.17  (0.4) &  2.81  (0.3) &  2.72  (0.6) \\
                                                      Develop health assessment methods or programs. &    2.18  (0.4) &  2.58  (0.5) &  2.37  (0.6) \\
                                                                 Administer intravenous medications. &    2.18  (0.4) &  2.76  (0.5) &  2.57  (0.6) \\
                        Collect information from people through observation, interviews, or surveys. &    2.18  (0.4) &  2.98  (0.4) &  1.73  (0.6) \\
                                             Develop treatment plans that use non-medical therapies. &    2.18  (0.4) &   2.0  (0.4) &  2.18  (0.7) \\
                                  Review professional literature to maintain professional knowledge. &    2.18  (0.4) &  3.17  (0.4) &  2.03  (0.6) \\
                                                Adjust tuning or functioning of musical instruments. &    2.18  (0.6) &  2.07  (1.1) &  2.81  (1.0) \\
                                                         Implement advanced life support techniques. &    2.19  (0.4) &  2.58  (0.4) &  2.71  (0.7) \\
                                                              Design medical devices or appliances. &    2.19  (0.5) &   2.3  (0.4) &  2.59  (0.7) \\
                Provide health and wellness advice to patients, program participants, or caregivers. &     2.2  (0.4) &  2.75  (0.4) &   2.6  (0.6) \\
                             Collaborate with healthcare professionals to plan or provide treatment. &    2.21  (0.4) &  2.81  (0.3) &  2.68  (0.6) \\
                                                         Maintain medical or professional knowledge. &    2.21  (0.4) &  2.96  (0.3) &  2.66  (0.6) \\
                                                                  Treat dental problems or diseases. &    2.21  (0.5) &  1.61  (0.4) &  2.73  (0.8) \\
                                                                      Write reports or evaluations. &    2.22  (0.4) &  3.71  (0.4) &  1.88  (0.6) \\
                                              Evaluate patient functioning, capabilities, or health. &    2.22  (0.4) &  2.52  (0.4) &  2.45  (0.6) \\
                                                             Administer non-intravenous medications. &    2.22  (0.4) &  3.28  (0.4) &  2.83  (0.6) \\
                                 Analyze patient data to determine patient needs or treatment goals. &    2.23  (0.4) &  2.65  (0.4) &  2.65  (0.6) \\
                                                                      Manage healthcare operations. &    2.23  (0.4) &  3.01  (0.4) &  2.57  (0.6) \\
                                              Encourage patients or clients to develop life skills. &    2.23  (0.5) &  1.98  (0.5) &  2.02  (0.7) \\
                                                                          Treat medical emergencies. &    2.23  (0.4) &  2.78  (0.4) &  2.77  (0.6) \\
                                                                      Administer cancer treatments. &    2.24  (0.5) &  2.71  (0.7) &  3.02  (0.7) \\
                                        Conduct research to increase knowledge about medical issues. &    2.24  (0.4) &  2.89  (0.3) &  2.61  (0.6) \\
                                          Examine mouth, teeth, gums, or related facial structures. &    2.24  (0.5) &  1.56  (0.4) &  3.06  (0.7) \\
                                                                    Teach health management classes. &    2.24  (0.5) &  2.54  (0.6) &  2.26  (0.6) \\
                          Explain medical procedures or test results to patients or family members. &    2.24  (0.4) &  2.97  (0.3) &  2.66  (0.6) \\
                                          Present social services program information to the public. &    2.25  (0.4) &  2.81  (0.4) &  2.03  (0.6) \\
                                                  Develop healthcare quality and safety procedures. &    2.25  (0.5) &  2.76  (0.6) &  2.96  (0.7) \\
                              Advise communities or institutions regarding health or safety issues. &    2.25  (0.4) &  3.05  (0.4) &  2.28  (0.6) \\
                                  Prepare reports summarizing patient diagnostic or care activities. &    2.25  (0.4) &  2.87  (0.4) &  2.59  (0.6) \\
                              Interact with patients to build rapport or provide emotional support. &    2.26  (0.4) &  2.91  (0.5) &  2.57  (0.7) \\
  
\bottomrule
\end{tabular}
\end{table}                                                       
                                                                
\newpage   
Appendix A Table (continued) 
\begin{table}[h!]
\centering
\small
\begin{tabular}{@{}p{8.5cm}ccc@{}}
\toprule
 \multicolumn{3}{r}{\textbf{Healthcare Experts}} \hspace{3em}  & \textbf{Tech} \\ 
\textbf{Work Activity} & \textbf{Automatability} &  \textbf{Desirability} & \textbf{Experts} \\
\midrule                                                            
                                                Treat patients using physical therapy techniques. &    2.27  (0.4) &  2.68  (0.6) &  2.81  (0.7) \\
                                          Communicate health and wellness information to the public. &    2.27  (0.4) &  2.87  (0.4) &  2.55  (0.6) \\
                                                         Determine protocols for medical procedures. &    2.27  (0.4) &  2.97  (0.4) &  3.08  (0.7) \\
                                                                  Record patient medical histories. &    2.27  (0.4) &   3.0  (0.3) &  2.77  (0.6) \\
                                                  Gather medical information from patient histories. &    2.28  (0.4) &  3.07  (0.4) &  2.72  (0.6) \\
                                                                  Present medical research reports. &    2.29  (0.4) &  2.83  (0.4) &  2.36  (0.6) \\
                                                Monitor patient progress or responses to treatments. &     2.3  (0.4) &   3.0  (0.3) &  2.57  (0.6) \\
                                                        Confer with clients to exchange information. &     2.3  (0.4) &  2.62  (0.4) &  2.24  (0.6) \\
                                                                            Train medical providers. &     2.3  (0.4) &  2.91  (0.3) &  2.79  (0.6) \\
                                                Treat patients using alternative medical procedures. &     2.3  (0.5) &  2.38  (0.8) &  2.91  (0.7) \\
                                 Evaluate patient outcomes to determine effectiveness of treatments. &     2.3  (0.4) &  2.82  (0.5) &  2.59  (0.6) \\                                 Test patient nervous system functioning. &    2.31  (0.4) &  2.42  (0.4) &  2.78  (0.6) \\
                                                            Care for patients with mental illnesses. &    2.32  (0.5) &  2.53  (0.9) &  2.24  (0.6) \\
                                          Prepare scientific or technical reports or presentations. &    2.32  (0.4) &  3.22  (0.4) &   2.2  (0.6) \\
                            Monitor patient conditions during treatments, procedures, or activities. &    2.32  (0.4) &  2.89  (0.3) &  2.74  (0.6) \\
                                                                  Maintain sterile operative fields. &    2.32  (0.5) &  2.72  (0.7) &  2.87  (0.7) \\
          Collect medical information from patients, family members, or other medical professionals. &    2.32  (0.4) &  3.14  (0.3) &  2.69  (0.6) \\
                                      Monitor the handling of hazardous materials or medical wastes. &    2.32  (0.5) &  2.87  (0.7) &  2.71  (0.7) \\
                                                             Test patient heart or lung functioning. &    2.32  (0.4) &  2.98  (0.4) &  2.69  (0.6) \\
                                    Communicate test or assessment results to medical professionals. &    2.34  (0.4) &  2.96  (0.4) &  3.06  (0.6) \\
 Train patients, family members, or caregivers in techniques for managing disabilities or illnesses. &    2.34  (0.4) &  2.93  (0.4) &   2.6  (0.6) \\
                                                                Order medical supplies or equipment. &    2.35  (0.4) &  3.11  (0.4) &  2.83  (0.6) \\
                                                                              Test patient hearing. &    2.35  (0.5) &  2.79  (1.0) &  2.74  (0.9) \\
                                                 Prepare patients physically for medical procedures. &    2.35  (0.4) &  2.46  (0.4) &  2.95  (0.6) \\
                                                      Adjust prostheses or other assistive devices. &    2.35  (0.5) &  2.81  (0.4) &  2.68  (0.7) \\
                                 Inform medical professionals regarding patient conditions and care. &    2.35  (0.4) &  3.23  (0.4) &  2.71  (0.6) \\
                                                  Advise others on business or operational matters. &    2.36  (0.5) &  2.92  (0.5) &  2.27  (0.6) \\
                                              Fit eyeglasses, contact lenses, or other vision aids. &    2.36  (0.5) &  2.58  (0.8) &  3.45  (0.7) \\
                          Test biological specimens to gather information about patient conditions. &    2.37  (0.4) &  3.04  (0.5) &  2.94  (0.6) \\
                                                              Supervise technical medical personnel. &    2.38  (0.5) &  2.75  (0.7) &  3.06  (0.7) \\
                                                                      Develop emergency procedures. &    2.38  (0.5) &  2.43  (0.8) &  2.32  (0.7) \\
                             Analyze laboratory specimens to detect abnormalities or other problems. &    2.38  (0.5) &  2.87  (0.6) &  3.05  (0.7) \\
                                          Follow protocols or regulations for healthcare activities. &    2.38  (0.4) &   2.7  (0.4) &  2.98  (0.6) \\
                                            Assist patients with hygiene or daily living activities. &    2.39  (0.5) &  3.14  (0.6) &  2.56  (0.6) \\
                              Consult with others regarding safe or healthy equipment or facilities. &    2.39  (0.5) &  2.43  (0.7) &  2.04  (0.7) \\
                                  Assist healthcare practitioners during examinations or treatments. &    2.39  (0.4) &  3.32  (0.4) &   2.9  (0.6) \\
                                                     Assist healthcare practitioners during surgery. &    2.39  (0.4) &  3.12  (0.4) &  2.89  (0.6) \\
                                Examine medical instruments or equipment to ensure proper operation. &    2.39  (0.4) &  2.93  (0.4) &   2.8  (0.6) \\
                                                      Prepare official health documents or records. &     2.4  (0.4) &  3.44  (0.4) &  2.78  (0.6) \\
                              Cultivate micro-organisms for study, testing, or medical preparations. &     2.4  (0.5) &  2.53  (0.8) &   2.9  (0.8) \\
                                 Operate diagnostic or therapeutic medical instruments or equipment. &     2.4  (0.4) &  3.19  (0.4) &  2.91  (0.6) \\
                                                    Train caregivers or other non-medical personnel. &     2.4  (0.4) &  2.67  (0.4) &  2.36  (0.6) \\
                                                      Manage preparation of special meals or diets. &     2.4  (0.5) &  1.94  (0.9) &  2.06  (0.7) \\
                                              Prepare biological specimens for laboratory analysis. &     2.4  (0.5) &  2.79  (0.8) &   3.2  (0.7) \\
                                                        Testify at legal or legislative proceedings. &     2.4  (0.5) &  2.89  (0.5) &  2.71  (0.6) \\
                                                         Conduct health or safety training programs. &    2.41  (0.5) &  2.69  (0.7) &  2.25  (0.6) \\
                                                      Prepare medical supplies or equipment for use. &    2.41  (0.4) &  3.14  (0.3) &  2.76  (0.6) \\
                                                              Apply bandages, dressings, or splints. &    2.41  (0.4) &  3.16  (0.5) &  2.78  (0.6) \\
                                                                          Fabricate medical devices. &    2.42  (0.5) &  2.91  (0.4) &  3.03  (0.7) \\
                                              Train personnel in technical or scientific procedures. &    2.42  (0.5) &   3.0  (0.4) &  2.55  (0.7) \\

\bottomrule
\end{tabular}
\end{table}                                                       
                                                                
\newpage   
Appendix A Table (continued) 
\begin{table}[h!]
\centering
\small
\begin{tabular}{@{}p{8.5cm}ccc@{}}
\toprule
 \multicolumn{3}{r}{\textbf{Healthcare Experts}} \hspace{3em}  & \textbf{Tech} \\ 
\textbf{Work Activity} & \textbf{Automatability} &  \textbf{Desirability} & \textbf{Experts} \\
\midrule                                                            

                                                             Verify accuracy of patient information. &    2.42  (0.4) &  3.34  (0.5) &  3.24  (0.6) \\
                                            Operate laboratory equipment to analyze medical samples. &    2.42  (0.5) &  2.84  (0.6) &  2.95  (0.7) \\
                                                              Prepare healthcare training materials. &    2.43  (0.4) &  2.96  (0.4) &  2.45  (0.6) \\
                                                     Position patients for treatment or examination. &    2.43  (0.4) &  3.08  (0.4) &  2.87  (0.6) \\
                                                          Test facilities for environmental hazards. &    2.43  (0.5) &  2.21  (0.9) &  2.19  (0.7) \\
                                                                      Process healthcare paperwork. &    2.43  (0.4) &  3.08  (0.5) &  2.78  (0.6) \\
                                                                                  Mediate disputes. &    2.43  (0.5) &  2.36  (0.5) &  2.26  (0.6) \\
                                      Administer medical substances for imaging or other procedures. &    2.43  (0.5) &  2.72  (0.5) &  2.94  (0.7) \\
                                                                        Analyze laboratory findings. &    2.44  (0.5) &  2.77  (0.9) &   3.0  (0.8) \\
                                          Protect patients or staff members using safety equipment. &    2.44  (0.5) &  2.97  (0.4) &  2.92  (0.6) \\
                                                         Inspect work environments to ensure safety. &    2.44  (0.5) &   2.4  (0.9) &  2.27  (0.7) \\
                                                              Recommend types of assistive devices. &    2.45  (0.5) &  3.37  (0.6) &  3.17  (0.6) \\
                                                                                Test patient vision. &    2.46  (0.5) &  3.43  (0.7) &  3.06  (0.7) \\
                                                                              Collect archival data. &    2.46  (0.5) &  3.55  (0.5) &  2.02  (0.7) \\
                                                              Operate diagnostic imaging equipment. &    2.46  (0.4) &  2.93  (0.4) &  2.95  (0.6) \\
                                                                  Maintain medical facility records. &    2.46  (0.4) &  3.38  (0.4) &  2.88  (0.6) \\
                                                          Prepare medications or medical solutions. &    2.47  (0.4) &  3.31  (0.5) &  3.02  (0.6) \\
                                                        Merchandise healthcare products or services. &    2.47  (0.5) &  3.17  (0.7) &  3.31  (0.7) \\
                                                         Sterilize medical equipment or instruments. &    2.47  (0.4) &  3.46  (0.4) &  2.98  (0.6) \\
                                                                 Check quality of diagnostic images. &    2.47  (0.5) &  2.89  (0.5) &  3.01  (0.7) \\
                                        Verify that medical activities or operations meet standards. &    2.47  (0.4) &  3.22  (0.4) &  2.95  (0.6) \\
                                                 Administer basic health care or medical treatments. &    2.47  (0.5) &  3.32  (0.6) &  2.69  (0.6) \\
                                                    Calculate numerical data for medical activities. &    2.47  (0.5) &  3.25  (0.5) &  2.93  (0.7) \\
                                       Measure the physical or physiological attributes of patients. &    2.48  (0.4) &  3.31  (0.5) &  2.84  (0.6) \\
                                                                  Repair medical facility equipment. &    2.48  (0.5) &  3.39  (0.4) &  3.04  (0.6) \\
                                                              Maintain medical laboratory equipment. &    2.48  (0.5) &  3.04  (0.7) &  3.06  (0.7) \\
                                                Maintain inventory of medical supplies or equipment. &    2.49  (0.4) &  3.52  (0.4) &  2.88  (0.6) \\
                                                              Clean medical equipment or facilities. &    2.49  (0.4) &  3.53  (0.4) &  2.82  (0.6) \\
                                                        Schedule patient procedures or appointments. &    2.49  (0.4) &  3.29  (0.4) &  2.96  (0.6) \\
                           Monitor video displays of medical equipment to ensure proper functioning. &    2.49  (0.5) &  2.91  (0.6) &  2.84  (0.6) \\
                          Create advanced digital images of patients using computer imaging systems. &     2.5  (0.5) &  3.02  (0.5) &   3.0  (0.6) \\
                                                                Record research or operational data. &     2.5  (0.5) &  2.93  (0.5) &   2.8  (0.7) \\
                                                  Adjust settings or positions of medical equipment. &     2.5  (0.4) &  3.22  (0.4) &  2.86  (0.6) \\
                                                           Move patients to or from treatment areas. &    2.52  (0.5) &  3.63  (0.5) &  2.71  (0.6) \\
                                                Instruct patients in the use of assistive equipment. &    2.52  (0.5) &  3.64  (0.6) &  2.95  (0.6) \\
                                                         Collect biological specimens from patients. &    2.53  (0.5) &  3.45  (0.5) &   2.9  (0.6) \\
                                                             Process x-rays or other medical images. &    2.53  (0.5) &  3.03  (0.5) &  2.94  (0.6) \\
                                                          Maintain medical equipment or instruments. &    2.54  (0.5) &  3.74  (0.4) &   2.9  (0.6) \\
                                                          Perform clerical work in medical settings. &    2.55  (0.5) &  3.25  (0.9) &  3.28  (0.6) \\
                Monitor medical facility activities to ensure adherence to standards or regulations. &    2.56  (0.5) &  2.93  (0.6) &   2.5  (0.6) \\
                                                                Supervise medical support personnel. &    2.56  (0.5) &  2.82  (0.6) &  2.72  (0.6) \\
                                                                Process medical billing information. &    2.57  (0.5) &  3.23  (0.8) &  3.31  (0.6) \\
                                                     Enter patient or treatment data into computers. &    2.57  (0.5) &  3.58  (0.5) &  3.21  (0.6) \\


\bottomrule
\end{tabular}
\end{table}

\newpage

\section*{Appendix B: Healthcare Occupations}

\begin{table}[h!]
\centering
\small
\caption{Healthcare occupations considered in our study. Those occupations who participated in the online survey and provided work activity labels are identified as Training Set (=1).}\label{tab:apps:occupations}
\begin{tabular}{@{}clc@{}}
\toprule
\textbf{SOC Code}   &  \textbf{Occupation Title}    &  \textbf{Training Set} \\
\midrule

29-1011.00 &                                      Chiropractors &               0 \\
    29-1021.00 &                                  Dentists, General &               1 \\
    29-1022.00 &                    Oral and Maxillofacial Surgeons &               1 \\
    29-1023.00 &                                      Orthodontists &               1 \\
    29-1024.00 &                                    Prosthodontists &               0 \\
    29-1029.00 &                    Dentists, All Other Specialists &               0 \\
    29-1031.00 &                       Dietitians and Nutritionists &               0 \\
    29-1041.00 &                                       Optometrists &               0 \\
    29-1051.00 &                                        Pharmacists &               0 \\
    29-1061.00 &                                  Anesthesiologists &               1 \\
    29-1062.00 &                   Family and General Practitioners &               1 \\
    29-1063.00 &                                Internists, General &               1 \\
    29-1064.00 &                    Obstetricians and Gynecologists &               1 \\
    29-1065.00 &                             Pediatricians, General &               1 \\
    29-1066.00 &                                      Psychiatrists &               1 \\
    29-1067.00 &                                           Surgeons &               1 \\
    29-1069.00 &                 Physicians and Surgeons, All Other &               0 \\
    29-1069.01 &                       Allergists and Immunologists &               0 \\
    29-1069.02 &                                     Dermatologists &               1 \\
    29-1069.03 &                                       Hospitalists &               0 \\
    29-1069.04 &                                       Neurologists &               1 \\
    29-1069.05 &                        Nuclear Medicine Physicians &               0 \\
    29-1069.06 &                                   Ophthalmologists &               0 \\
    29-1069.07 &                                       Pathologists &               0 \\
    29-1069.08 &    Physical Medicine and Rehabilitation Physicians &               0 \\
    29-1069.09 &                     Preventive Medicine Physicians &               0 \\
    29-1069.10 &                                       Radiologists &               0 \\
    29-1069.11 &                         Sports Medicine Physicians &               0 \\
    29-1069.12 &                                         Urologists &               0 \\
    29-1071.00 &                               Physician Assistants &               0 \\
    29-1071.01 &                        Anesthesiologist Assistants &               0 \\
    29-1081.00 &                                        Podiatrists &               0 \\
    29-1122.00 &                            Occupational Therapists &               1 \\
    29-1122.01 &  Low Vision Therapists, Orientation and Mobilit... &               0 \\
    29-1123.00 &                                Physical Therapists &               0 \\
    29-1124.00 &                               Radiation Therapists &               0 \\
    29-1125.00 &                            Recreational Therapists &               0 \\
    29-1125.01 &                                     Art Therapists &               0 \\
    29-1125.02 &                                   Music Therapists &               0 \\
    29-1126.00 &                             Respiratory Therapists &               1 \\
    29-1127.00 &                       Speech-Language Pathologists &               0 \\
    29-1128.00 &                             Exercise Physiologists &               0 \\
    29-1129.00 &                              Therapists, All Other &               0 \\
    29-1131.00 &                                      Veterinarians &               0 \\
    29-1141.00 &                                  Registered Nurses &               0 \\
    29-1141.01 &                                  Acute Care Nurses &               0 \\
    29-1141.02 &               Advanced Practice Psychiatric Nurses &               0 \\
    29-1141.03 &                               Critical Care Nurses &               0 \\
    29-1141.04 &                         Clinical Nurse Specialists &               0 \\
    29-1151.00 &                                 Nurse Anesthetists &               0 \\
    29-1161.00 &                                     Nurse Midwives &               0 \\
    29-1171.00 &                                Nurse Practitioners &               0 \\
    29-1181.00 &                                       Audiologists &               0 \\
    29-1199.00 &  Health Diagnosing and Treating Practitioners, ... &               0 \\
    29-1199.01 &                                     Acupuncturists &               0 \\
\bottomrule
\end{tabular}
\end{table}

\newpage

\begin{table}[h!]
\centering
\small
\begin{tabular}{@{}clc@{}}
\toprule
\textbf{SOC Code}   &  \textbf{Occupation Title}    &  \textbf{Training Set} \\
\midrule
29-1199.04 &                            Naturopathic Physicians &               0 \\
    29-1199.05 &                                        Orthoptists &               0 \\
    29-2011.00 &      Medical and Clinical Laboratory Technologists &               0 \\
    29-2011.01 &                          Cytogenetic Technologists &               0 \\
    29-2011.02 &                                  Cytotechnologists &               0 \\
    29-2011.03 &      Histotechnologists and Histologic Technicians &               0 \\
    29-2012.00 &        Medical and Clinical Laboratory Technicians &               0 \\
    29-2021.00 &                                  Dental Hygienists &               0 \\
    29-2031.00 &       Cardiovascular Technologists and Technicians &               0 \\
    29-2032.00 &                    Diagnostic Medical Sonographers &               0 \\
    29-2033.00 &                     Nuclear Medicine Technologists &               0 \\
    29-2034.00 &                           Radiologic Technologists &               0 \\
    29-2035.00 &           Magnetic Resonance Imaging Technologists &               0 \\
    29-2041.00 &       Emergency Medical Technicians and Paramedics &               0 \\
    29-2051.00 &                               Dietetic Technicians &               0 \\
    29-2052.00 &                               Pharmacy Technicians &               0 \\
    29-2053.00 &                            Psychiatric Technicians &               0 \\
    29-2054.00 &                    Respiratory Therapy Technicians &               0 \\
    29-2055.00 &                             Surgical Technologists &               0 \\
    29-2056.00 &           Veterinary Technologists and Technicians &               0 \\
    29-2057.00 &                     Ophthalmic Medical Technicians &               0 \\
    29-2061.00 &  Licensed Practical and Licensed Vocational Nurses &               0 \\
    29-2071.00 &  Medical Records and Health Information Technic... &               0 \\
    29-2081.00 &                              Opticians, Dispensing &               0 \\
    29-2091.00 &                        Orthotists and Prosthetists &               0 \\
    29-2092.00 &                            Hearing Aid Specialists &               0 \\
    29-2099.00 &    Health Technologists and Technicians, All Other &               0 \\
    29-2099.01 &                      Neurodiagnostic Technologists &               0 \\
    29-2099.05 &                   Ophthalmic Medical Technologists &               0 \\
    29-2099.06 &                             Radiologic Technicians &               0 \\
    29-2099.07 &                                Surgical Assistants &               0 \\
    29-9011.00 &         Occupational Health and Safety Specialists &               0 \\
    29-9012.00 &         Occupational Health and Safety Technicians &               0 \\
    29-9091.00 &                                  Athletic Trainers &               0 \\
    29-9092.00 &                                 Genetic Counselors &               0 \\
    29-9099.00 &  Healthcare Practitioners and Technical Workers... &               0 \\
    29-9099.01 &                                           Midwives &               0 \\
    19-3031.02 &                             Clinical Psychologists &               1 \\
    19-3031.03 &                           Counseling Psychologists &               1 \\
    19-3032.00 &            Industrial-Organizational Psychologists &               1 \\
    19-3039.01 &  Neuropsychologists and Clinical Neuropsycholog... &               1 \\
    21-1013.00 &                     Marriage and Family Therapists &               1 \\
\bottomrule
 & & \\
 & & \\
 & & \\
 & & \\
 & & \\
 & & \\
 & & \\
 & & \\
 & & \\
 & & \\
 & & \\
 & & \\
 & & \\
 & & \\
 & & \\
 & & \\
 & & \\
& & \\
& & \\
\end{tabular}
\end{table}

\section*{Appendix C: Surveyed Healthcare Occupation}

\begin{table}[h!]
\centering
\small
\caption{Work activities for two healthcare occupations that participated in our online expert survey.}\label{tab:apps:activities}
\begin{tabular}{@{}lcl@{}}
\toprule
\textbf{SOC Code/Occupation} &  \textbf{Work Activity ID} &  \textbf{Work Activity} \\
\midrule
29-1067.00 & 4.A.4.c.3.I06.D01 &                                            Prescribe treatments or therapies. \\
Surgeons & 4.A.4.c.3.I06.D03 &                                                        Prescribe medications. \\
& 4.A.2.a.3.I04.D02 &                    Follow protocols or regulations for healthcare activities. \\
& 4.A.3.b.6.I11.D04 &                                             Record patient medical histories. \\
& 4.A.1.b.2.I09.D03 &                        Examine patients to assess general physical condition. \\
& 4.A.2.b.1.I06.D02 &                                                  Diagnose medical conditions. \\
& 4.A.4.a.5.I01.D02 &         Refer patients to other healthcare practitioners or health resources. \\
& 4.A.2.b.5.I03.D02 &                                  Schedule patient procedures or appointments. \\
& 4.A.4.a.5.I12.D03 &                                      Operate on patients to treat conditions. \\
& 4.A.4.b.4.I01.D14 &                                             Supervise patient care personnel. \\
& 4.A.2.a.4.I06.D02 &           Analyze patient data to determine patient needs or treatment goals. \\
& 4.A.1.a.1.I19.D01 &                  Conduct research to increase knowledge about medical issues. \\
& 4.A.4.c.3.I05.D07 &                                          Order medical supplies or equipment. \\
& 4.A.4.b.4.I12.D38 &                                                 Manage healthcare operations. \\
& 4.A.4.b.6.I06.D01 &                         Advise medical personnel regarding healthcare issues. \\
& 4.A.4.a.5.I11.D06 &                               Assist healthcare practitioners during surgery. \\
& 4.A.3.a.1.I09.D01 &                                   Sterilize medical equipment or instruments. \\ 
\cdashline{1-3}
19-3031.02 & 4.A.3.b.6.I03.D04 &                     Prepare scientific or technical reports or presentations. \\

Clinical Psychologists & 4.A.2.b.3.I01.D01 &            Review professional literature to maintain professional knowledge. \\
& 4.A.4.a.2.I13.D02 &                                  Confer with clients to exchange information. \\
& 4.A.4.b.4.I06.D09 &                                Direct medical science or healthcare programs. \\
& 4.A.2.b.2.I15.D08 &                                                 Develop educational programs. \\
& 4.A.1.a.1.I12.D03 &  Collect information from people through observation, interviews, or surveys. \\
& 4.A.4.b.6.I06.D05 &                                          Advise others on healthcare matters. \\
& 4.A.4.b.4.I01.D03 &                                                           Supervise trainees. \\
& 4.A.2.b.2.I23.D08 &                                                Plan social sciences research. \\
& 4.A.4.c.2.I02.D03 &                      Administer standardized physical or psychological tests. \\
& 4.A.4.b.6.I10.D06 &                     Counsel clients on mental health or personal achievement. \\
& 4.A.2.b.2.I15.D02 &         Design psychological or educational treatment procedures or programs. \\
& 4.A.2.b.1.I06.D03 &                                   Diagnose neural or psychological disorders. \\
& 4.A.1.a.1.I04.D08 &                                                        Collect archival data. \\
\bottomrule
\end{tabular}
\end{table}

\end{document}